\pdfoutput=1

\documentclass[11pt]{article}

\usepackage[final]{EMNLP2022}

\usepackage{times}
\usepackage{latexsym}

\usepackage{microtype}

\usepackage[T1]{fontenc}

\usepackage[utf8]{inputenc}

\usepackage{microtype}
\usepackage{graphicx}
\usepackage{amsmath}

\usepackage[export]{adjustbox}
\usepackage{array}

\usepackage{enumitem}

\usepackage[ruled,vlined]{algorithm2e}
\DeclareMathAlphabet{\pazocal}{OMS}{zplm}{m}{n}

\usepackage{graphicx}               
\usepackage{tabularx}               
\newcolumntype{C}{>{\centering\arraybackslash}X}
\usepackage{multirow}               
\usepackage{diagbox}                
\usepackage{hhline}                 
\usepackage{color}                  
\usepackage{amssymb}                
\usepackage{mathtools}              
\usepackage{enumitem}               

\usepackage{makecell}
\usepackage{xcolor}

\DeclareMathOperator*{\argmax}{arg\,max}

\definecolor{effectspancolor}{RGB}{0, 51, 125}
\definecolor{drugspancolor}{RGB}{44, 7, 110}

\usepackage{xparse}
\NewDocumentCommand{\heng}{ mO{} }{\textcolor{red}{\textsuperscript{\textit{Heng}}\textsf{\textbf{\small[#1]}}}}
\NewDocumentCommand{\tuan}{ mO{} }{\textcolor{blue}{\textsuperscript{\textit{Tuan}}\textsf{\textbf{\small[#1]}}}}
\NewDocumentCommand{\zixuan}{ mO{} }{\textcolor{orange}{\textsuperscript{\textit{Zixuan}}\textsf{\textbf{\small[#1]}}}}
\NewDocumentCommand{\han}{ mO{} }{\textcolor{blue}{\textsuperscript{\textit{Han}}\textsf{\textbf{\small[#1]}}}}
\NewDocumentCommand{\clare}{ mO{} }{\textcolor{orange}{\textsuperscript{\textit{Clare}}\textsf{\textbf{\small[#1]}}}}
\NewDocumentCommand{\liliang}{ mO{} }{\textcolor{brown}{\textsuperscript{\textit{Liliang}}\textsf{\textbf{\small[#1]}}}}
\NewDocumentCommand{\cheng}{ mO{} }{\textcolor{purple}{\textsuperscript{\textit{Cheng}}\textsf{\textbf{\small[#1]}}}}

\usepackage{microtype}
\usepackage[utf8]{inputenc} 
\usepackage[T1]{fontenc}    
\usepackage{hyperref}       
\usepackage{graphicx}
\usepackage{url}            
\usepackage{booktabs}       
\usepackage{amsfonts}       
\usepackage{nicefrac}       
\usepackage{xcolor}
\usepackage{bm}
\usepackage{multirow}
\usepackage{makecell}
\usepackage{amsthm}
\usepackage{cleveref}

\usepackage{inconsolata}

\title{Language Model Pre-Training with Sparse Latent Typing}

\author{
  Liliang Ren$^{1*}$, Zixuan Zhang$^{1}$\thanks{\; Equal contribution. 
Listing order is random. 
Liliang proposed and implemented the architecture designs and the training objectives of Sparse Latent Typing (SLT), and he also conducted extensive experiments for pre-training, few-shot evaluation and the analyses. 
Zixuan designed the language model pre-training pipeline for SLT, built the initial training codebase and conducted the experiments for pre-training and supervised evaluation.
Both of the authors initially came up with the same project goal of encouraging the model to sparsely select sentence-level key words during pre-training.
  }, \textbf{Han Wang}$^{2}$\textbf{,}  \textbf{Clare R. Voss}$^{3}$\textbf{,} \textbf{Chengxiang Zhai}$^{1}$\textbf{,} \textbf{Heng Ji}$^{1}$ \\
  $^1$University of Illinois at Urbana-Champaign, 
  $^2$Amazon Alexa,\\ $^3$US Army Research Laboratory \\
  \texttt{\{liliang3, zixuan11, czhai, hengji\}@illinois.edu} \\
  \texttt{wnghn@amazon.com}, ~\texttt{clare.r.voss.civ@army.mil}
  }

\begin{document}
\maketitle
\begin{abstract}
    Modern large-scale Pre-trained Language Models (PLMs) have achieved tremendous success on a wide range of downstream tasks. However, most of the LM pre-training objectives only focus on text reconstruction, but have not sought to learn latent-level interpretable representations of sentences.
    In this paper, we manage to push the language models to obtain a deeper understanding of sentences by proposing a new pre-training objective, \emph{Sparse Latent Typing}, which enables the model to sparsely extract sentence-level keywords with diverse latent types.
    Experimental results show that our model is able to learn interpretable latent type categories in a self-supervised manner without using any external knowledge.
    Besides, the language model pre-trained with such an objective also significantly improves Information Extraction related downstream tasks in both supervised and few-shot settings. Our code is publicly available at \url{https://github.com/renll/SparseLT}.
\end{abstract}

\section{Introduction}
Transformer-based Pre-trained Language Models (PLMs) have achieved significant success on a wide range of NLP tasks.
However, typical pre-training objectives for PLMs only focus on teaching the model to directly reconstruct text-level words or sentences, but have not sought to obtain deeper sentence understanding by learning latent-level interpretable representations.
For example, transformer-decoder models like the OpenAI GPT series~\cite{gpt, gpt2, gpt3} adopt the task of language modeling for pre-training, and transformer-encoder models like BERT~\cite{bert} and RoBERTa~\cite{roberta} are trained by predicting the masked tokens within a sentence.
Both of these training objectives merely train the models to recover the masked tokens or predict the next words or sentences, while ignoring to learn latent-level representations of sentences that could be potentially useful for both better language understanding and downstream tasks.



Pre-training a language model to learn latent representations is extremely hard:
First, there are no ground-truth labels for the latent representations that could be used for reliable supervised learning.
During pre-training, the model is only given an unlabeled text corpus over which to identify latent representations such as sentence-level keywords and structures. This means the training process must be strictly self-supervised~\cite{rush-etal-2018-deep}.
Furthermore, to be interpretable, the latent representations for natural language texts are supposed to be discrete, which further complicates the design of a completely differentiable training framework. 
\begin{figure}[tb]
  \centering
  \includegraphics[width=0.49\textwidth]{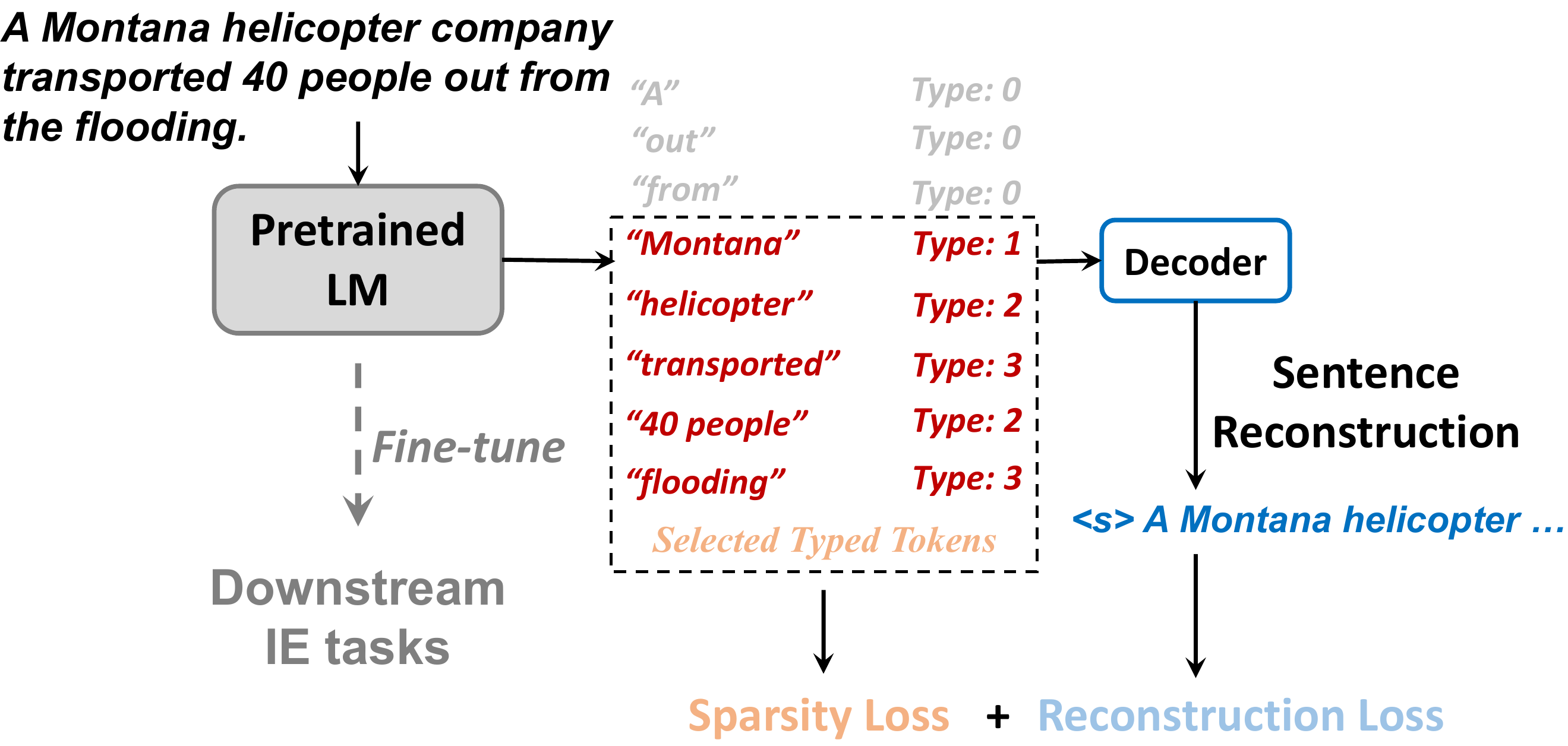}
  \caption{A general illustration of our approach to teach pre-trained language model to extract sentence-level keywords with latent type representations in a completely self-supervised manner.}\label{fig:framework}
  \vspace{-0.15cm}
\end{figure}

To push the language models to learn deeper understandings of sentences, in this paper, we propose a novel pre-training framework, \emph{Sparse Latent Typing}, that enables the language model to sparsely extract sentence-level keywords with meaningful latent types.
We have tackled all above-mentioned challenges and our framework is fully differentiable and completely self-supervised.
As shown in Figure~\ref{fig:framework}, given an input sentence from the pre-training corpus, we introduce a latent typing mechanism to jointly selects and classifies the keywords from the sentence into a category of randomly initialized latent types.
We implement such an latent classification model based on Gumbel Sampling~\cite{gumbel} to make sure the overall pre-training framework is differentiable.
Since there are no ground-truth labels available for the selected keywords and latent types, we incorporate an one-layer transformer decoder into the training pipeline to map the fused token and latent type representations back to the original sentence, and use the sentence reconstruction loss to control for adequate usefulness of the latent representations. Our approach provides the decoder model with a shortcut to directly access the encoded token representations, so that the latent representation for each of the input tokens can be learned as an auxiliary type representation.
For pre-training objectives, in addition to minimizing the sentence reconstruction error, we also introduce a novel typing sparsity loss to minimize the number of token representation selected for latent typing. A KL-divergence based diversity loss is also proposed to encourage a diverse selection of the latent types.
Experimental results show that our model is able to learn interpretable latent type categories in a self-supervised manner without using any external knowledge.
Besides, the language model pre-trained with such an objective also significantly improves Information Extraction related downstream tasks in both supervised and few-shot settings.

In summary, our contributions are three-fold:
\begin{itemize}
    \item We propose a fully differentiable language model pre-training framework that enables the model to sparsely extract sentence-level keywords with latent types in a completely self-supervised manner.
    
    \item We provide comprehensive analysis and interpretation for our experimental results showing that the pre-trained model is able to extract meaningful latent type representations.
    
    
    \item Extensive experiments on IE-related downstream tasks demonstrate that our proposed pre-training framework can significantly advance state-of-the-art. 
    
\end{itemize}

\section{Related Work}
\paragraph{Knowledge-Enhanced Language Models}
As pretrained language models \cite{gpt, bert, roberta, gpt2, gpt3, bart, t5} are achieving great success on downstream NLP tasks, many research studies focus on how to make these PLMs more knowledgeable.
Previous studies~\cite{knowbert,ernie,encycopedia,bertmk,luke,erica,kepler} either focus on designing entity-relation-aware pre-training objectives, or modifying the model architecture to make it capable of fusing both text and entity information.
However, all of these previous approaches utilize large-scale, human-annotated, semi-structured external resources (e.g., Wikipedia).
In comparison, our method is completely self-supervised and only needs a text corpus for pre-training, which focuses more on encouraging the model to learn knowledge clusters at a latent level.
\paragraph{Latent Structure Learning}
There are also several studies~\cite{liu-etal-2021-awakening,subramani-etal-2022-extracting} that incorporate latent structure learning into language model pre-training.
Particularly, \newcite{sentenceae} also proposes to use a transformer decoder layer to reconstruct the original sentence to provide training signals.
However, instead of learning coarse-grained sentence representations, we focus on learning fine-grained latent type representation that are interpretable and useful at the token level. To meet this end, we propose a series of novel training objectives and architecture designs to facilitate a sparse selection and typing of the token representations in the latent space.



\paragraph{Information Extraction}
Our approach to detect sentence-level keywords with latent types is inspired by Information Extraction (IE) \cite{ie}, an essential NLP task that aims to extract knowledge from texts.
Although IE includes a wide range of tasks varying in \emph{what to extract} (entities, relations, events) and \emph{where to extract from} (sentences, documents, corpora), typical IE frameworks usually include two essential steps: 1) \emph{Selection}: selecting the most task-relevant units from the inputs, 2) \emph{Classification}: assigning each of these a correct type label.
Such a select-and-classify framework is common to several IE tasks, including entity extraction, event detection and event argument extraction.
Accordingly, in our approach, we follows a similar \emph{Selection-Classification} approach to incorporate word selections and latent typing in pre-training.

\section{Problem Formulation}

\begin{figure*}
  \centering
  \includegraphics[width=16cm]{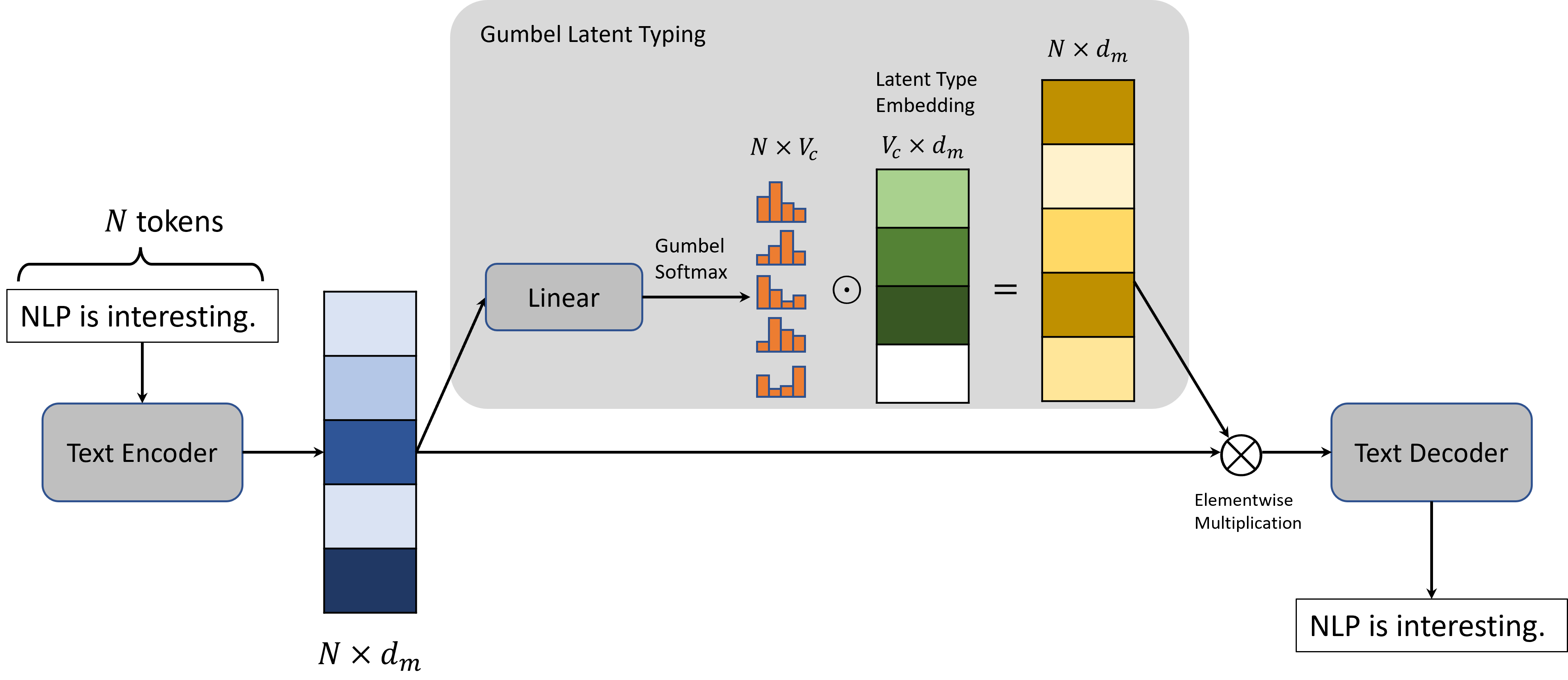}
  \caption{The proposed architecture for pre-training a language model with Gumbel Latent Typing, where $d_m$ is the length of the token representation vectors, $V_c$ is the pre-defined size of the latent types, and $\odot$ means the matrix multiplication. The white block in the latent type embedding is the zero type vector.}\label{arch}
  \vspace{-0.1cm}
\end{figure*}

Given a text corpus $\mathcal{D}$ composed of text sentences $\mathcal{S}$, we use $\mathbf{s}=\{w_1, \cdots,w_N\}, \mathbf{s} \sim \mathcal{S}$ to represent a sentence consisting of $N$ tokens. Assuming a text encoder $f : \mathcal{S} \mapsto \mathcal{X}$ that takes the sentence $\mathbf{s}$  as input and outputs the token representations $\mathbf{x}_1, \cdots, \mathbf{x}_N$, our latent type classifier $h: \mathcal{X} \mapsto (\mathcal{Z},\hat{{\cal X}}) $ then selects a subset of token representations $\hat{\mathbf{x}}_1, \cdots, \hat{\mathbf{x}}_T, T\leq N $ and classifies them into latent type representations $\mathbf{z}_1, \cdots, \mathbf{z}_T $. Each of the token types $\mathbf{z}_i$ is selected from a latent embedding space $\mathcal{C}$ consisting of $ V_c = |\mathcal{C}|$ different latent vectors.
The text decoder $g: (\mathcal{Z},\hat{{\cal X}}) \mapsto \mathcal{S} $ then reconstructs the original sentence $\mathbf{s}$ through the pair of latent types and selected token representations $(\mathcal{Z},\hat{{\cal X}})$.

The objective of sparse latent typing is to find pairs of latent types and token representations that are as compact as possible but still contain the necessary information for reconstructing the original input sentences. Formally, we want to minimize the following joint objective, 
\begin{equation} \label{obj}
    \min_{\theta_f,\theta_h, \theta_g} ~ T, \mathcal{L}_{\text{rec}}(f, h, g), \mathcal{L}_{\text{KL}}(f,h)
\end{equation}
with:
\begin{align*}
    \mathcal{L}_{\text{rec}}(f, h, g) &= ~\mathbb{E}_{~\mathbf{s}  \sim \mathcal{S}}[-\log p_g(\mathbf{s}|h(f(\mathbf{s})))],\\
     \mathcal{L}_{\text{KL}}(f,h) &= D_{\text{KL}}(p_h(\mathbf{z}|f(\mathbf{s}))||p(\mathbf{z})),
\end{align*}
where $T$ is the number of the selected token representations, $p(\mathbf{z})$ is a prior distribution of the latent types, and $D_{\text{KL}}(\cdot || \cdot )$ is the Kullback–Leibler (KL) divergence. The reconstruction loss and the KL term in our formulation follows the classical VAE \cite{vae}, but there are two key differences: (1) The latent variables $\mathbf{z}$ are discrete categorical variables, (2) Instead of only taking the latent representation $z$, the decoder takes both the token vectors and the corresponding latent vectors for sentence reconstruction.   Since the discrete version of VAE is well studied by the previous efforts such as VQ-VAE \cite{vqvae} and Gumbel-Softmax \cite{gumbel}, the optimization problem remains as how to minimize the non-differentiable term $T$ to encourage the sparse selection of the token representations.

\section{Learning Sparse Latent Types}

To tackle the non-differentiable problem of the size of the selected typing pairs $T = |(\mathcal{Z},\hat{{\cal X}})|$, we first take a closer look at the latent type classifier $h$ which decides the latent type $\mathbf{z}_i$ of each token representation $\mathbf{x}_i$. Our insight is that we can regard the action of not selecting a token representation as a frozen zero type vector $\mathbf{c}_1= \mathbf{0} \in \mathcal{C}$. We then do an element-wise multiplication between $\mathbf{z}_i$ and $\mathbf{x}_i$ to obtain the representations $\bar{\bf x}_i = {\bf x}_i \otimes {\bf z}_i$ that are to be fed into the text decoder $g$. The advantages of this approach are that (1) the element-wise multiplication naturally prevents the gradient from being propagated to the token representations that are classified as the zero type vector $\mathbf{c}_1$, (2) the element-wise multiplication directly modulates the gradients of the token representations with the latent type vectors. This can in principle provide better guidance to the text encoder with the information of the latent vectors than can be provided by other vector fusion operators such as element-wise addition or vector concatenation. Based on this framework, we developed a novel typing sparsity loss in \Cref{to} to approximately minimize the typing pairs size $T$. While our approach is generally applicable for any text encoder and decoder, specific neural architectures used in this work are discussed in \Cref{slt}.

In our framework, the latent type classifier $h$ is simplified as a mapping $h': \mathcal{X} \mapsto \mathcal{Z}$ that only outputs the latent types ${\bf z}_i$ for each token representation. The simplified text decoder $g': \bar{{\cal X}} \mapsto \mathcal{S} $ then only needs to model the fused representation space $\bar{{\cal X}} =  \cal Z \otimes \cal X $ for sentence reconstruction. $\otimes$ is the vector fusion operator and should be interpreted as element-wise multiplication in this work. The proposed architecture for sparse latent typing is illustrated in \Cref{arch}, which is further explained in the following subsections.

\subsection{Gumbel Latent Typing}

Given the token representations generated from the text encoder, $X=\{ {\bf x}_1, \cdots, {\bf x}_N \} \in \mathbb{R}^{N\times d_m}$, where $N$ is the number of input tokens, and $d_m$ is the length of the token representation vectors, our Gumbel latent type classifier first maps $X$ into logits ${L} \in \mathbb{R}^{N\times V_c}$  with a weight matrix $W \in \mathbb{R}^{d_m \times V_c}$, and then outputs the probabilities $P_{i,v}$ of choosing the $v$-th latent type for each token representation $\mathbf{x}_i$,
\begin{align*}
    P_{i,v} &= \frac{e^{(L_{i,v}+G_{i,v})/\tau}}{\sum_{k=1}^{V_c} e^{(L_{i,k}+G_{i,k})/\tau}},\\
    L &= XW,
\end{align*}
where $G_{i,v} \sim \text{Gumbel}(0,1)$ is the Gumbel noise sampled from a standard Gumbel distribution and $\tau$ is the non-negative temperature, following the previous efforts on  the Gumbel softmax operation \cite{gumbel, concrete}. The reason why we are using Gumbel softmax for our latent type classifier is that it enables choosing a latent type representation in a fully differentiable way, and thus can further facilitate our design of the sparsity loss to do an approximate minimization of the size of the typing pairs $T$.

With the Gumbel decision probability $P\in \mathbb{R}^{N \times V_c}$, the latent type representations $Z \in \mathbb{R}^{N\times d_m} $ are obtained through a marginalization of $P$ over the latent type embeddings $C \in \mathbb{R}^{V_c \times d_m}$,
\begin{align*}
    \mathbf{c}_1 &= \mathbf{0},\\
    Z &= \{\mathbf{z}_1, \cdots, \mathbf{z}_N \}, \\
    \mathbf{z}_i &= \sum_{k=1}^{V_c} P_{i,k}\mathbf{c}_k,
\end{align*}
where $\mathbf{c}_1$ is the zero type vector. The final fused representation $\bar{X}$ is obtained by an element-wise multiplication,
$$
\bar{X} = X\otimes Z.
$$
Intuitively, if $P_{i,k}$ is entirely concentrated on $c_1$ as $\tau \rightarrow 0$, i.e., $P_{i,1}=1$, we effectively eliminate token $w_i$ (or its representation $x_i$).

During the evaluation stage of our latent type classifier, the latent type embedding with the largest logit score is selected as the latent type representation for each of the token vectors $\mathbf{x}_i$,
\begin{align*}
    k^* &= \argmax_k L_{i,k},\\
    \mathbf{z}_i &= \mathbf{c}_{k^*}.
\end{align*}
To alleviate the discrepancy between the training and the evaluation, we adopt the temperature annealing \cite{gumbel} trick to the Gumbel-Softmax for a better differentiable approximation of the argmax operator.

\subsection{Training Objectives} \label{to}

Based on our problem formulation, we adopt three types of training loss for the end-to-end training of our model: (1) \emph{Typing Sparsity Loss} that encourages the latent type classifier to choose more zero types,
(2) \emph{KL-Divergence} with respect to a uniform prior distribution to encourage the diverse selection of the latent types,
(3) \emph{Reconstruction Loss} that ensures the latent representation maintains essential information of the input text.
\paragraph{Typing Sparsity Loss} An important property of Gumbel-Softmax is that when the temperature $\tau \rightarrow 0$, the decision probability $P_i \in \mathbb{R}^{V_c} $ will tend to be an one-hot index vector sampled from the underlying categorical distribution,
\begin{align*}
    \hat{P}_i(C = v) = \frac{e^{L_{i,v}}}{\sum_{k=1}^{V_c} e^{L_{i,k}}},
\end{align*}
where $L\in \mathbb{R}^{N\times V_c}$ is the logits before doing Gumbel-Softmax. This means that we can control the decision behavior of the model through modulating the shape of this categorical distribution. Therefore, our typing sparsity loss is designed as the negative log-likelihood of the global averaged probability of choosing the zero type $\mathbf{c}_1$,

$$
\mathcal{L}_s = -\log \frac{1}{N} \sum_{i=1}^N{\hat{P}_i(C= 1 )},
$$
where $N$ is the number of tokens in the input sentence. Intuitively, if $\hat{P}_i$ converges to an one-hot vector, then $\hat{P}_i(C= 1 ) \in \{1,0\} $, and $\sum_{i=1}^N{\hat{P}_i(C= 1 )} = N-T$ becomes the number of the tokens that are not selected for typing, which is equivalent to what we want to maximize in the problem formulation of \Cref{obj}.

\paragraph{KL-Divergence} To encourage a diverse selection of the latent types, we assume a uniform prior distribution of the latent type representations $p(\mathbf{z}) = 1/V_c$. The KL-divergence term is calculated between the global averaged probability $\bar{P}_v$ and the uniform prior, \emph{i.e.},
\begin{align*}
   \mathcal{L}_{\text{KL}}&= \frac{1}{V_c}\sum_{v=1}^{V_c}\bar{P}_v\log (\bar{P}_v ),\\
   \bar{P}_v &=\frac{1}{N} \sum_{i=1}^N\hat{P}_{i,v},
\end{align*}
where $V_c$ is the number of the latent types.

\paragraph{Reconstruction Loss} Our reconstruction loss directly follows our problem formulation, \emph{i.e.},
$$
\mathcal{L}_{\text{rec}} = ~ -\frac{1}{B}\sum_{i=1}^B\log p_g(\mathbf{s}_i|h(f(\mathbf{s}_i))),
$$
where $B$ is the batch size of the sampled text data. When pre-training a masked language model, we also include a Masked Language Model loss following BERT \cite{bert},
\begin{equation}\label{eqn:mlm_loss}
\mathcal{L}_{\text{MLM}} = ~ -\frac{1}{B}\sum_{i=1}^B\log p_f(\mathbf{s}_i|\tilde{\mathbf{s}}_i),
\end{equation}
where $f$ is the text encoder, and $\tilde{\mathbf{s}}_i$ is the corrupted sequence. 

The total loss function is a weighted sum of the above four losses,
\begin{equation}\label{eqn:final_loss}
\mathcal{L} = \mathcal{L}_{\text{MLM}} + \alpha \mathcal{L}_{\text{rec}} + \beta \mathcal{L}_s +\gamma \mathcal{L}_{\text{KL}},
\end{equation}
where $\alpha, \beta, \gamma \in \mathbb{R}_{\geq 0 }$ are weighting factors.

\section{Experiments}
In our experiments, we first conduct intrinsic evaluation to investigate whether the model can successfully learn word selections with meaningful latent types during pre-training.
Then, we apply our model on both supervised and few-shot IE tasks to evaluate the effectiveness of our pre-training framework on downstream tasks.

\subsection{Sparse Latent Type Learning}\label{slt}
\paragraph{Pre-training Setup}
We adopt the VOA corpus constructed by~\cite{m2e2} for sparse latent type pre-training, which was extracted from 108,693 multimedia news articles openly available on the Voice of America website between 2006 and 2017.
We use the \emph{bert-base-uncased} version of the BERT~\cite{bert} model as our encoder, and a single transformer decoder layer to reconstruct the sentence, following \citet{kasai2020deep, sentenceae}. While our approach is generally applicable for both encoder-only Masked Language Model (MLM) and the encoder-decoder Denoising Language Model (e.g. BART \cite{lewis-etal-2020-bart}), we focus on MLM because MLM is more widely used in the downstream information extraction tasks. 
The implementation details can be found in Appendix~\ref{appx:implementation}. 


\paragraph{Downstream Evaluation}
To 
evaluate the validity of our latent typing approach, we apply our pre-trained model by fine-tuning on downstream tasks.
We focus on Information Extraction specifically, and adopt \emph{Supervised Joint Information Extraction}~\cite{oneie} and \emph{Few-shot Named Entity Recognition}~\cite{ding2021few} as two typical IE tasks to evaluate our model on both supervised and few-shot IE settings.
We initialize the BERT model with \emph{bert-base-uncased} weights and continue pre-training 100,000 steps using the combined loss defined in \eqref{eqn:final_loss} on the VOA corpus. Since we only focus on evaluating our model on the IE tasks, only the pre-trained text-encoder is used for fine-tuning on the downstream tasks.
More details can be found in Appendix~\ref{appx:implementation}.

\subsection{Supervised Information Extraction}

\paragraph{Datasets}
We evaluate our pre-trained model on the English subset of ACE-2005 dataset\footnote{\url{https://catalog.ldc.upenn.edu/LDC2006T06}} and the ERE dataset, which are the most widely-used event-centric IE dataset containing annotations for extracting entities, events, and relations.
Following the preprocessing steps and dataset splits in \cite{oneie}, we keep 7 entity types, 6 relation types, 33 event types, and 22 argument roles for the ACE-2005 dataset, and 7 entity types, 5 relation types, 38 event types, and 20 argument roles for the ERE dataset.
More detailed dataset statistics are shown in Table~\ref{tab:supervised_dataset_stats}.
\begin{table}[htbp]
	\centering
    \small
	\begin{tabular}{cccccc}
		\toprule[1pt] 
		Dataset & Split & \#Sents & \#Ents & \#Events & \#Rels\\
		\midrule[1pt]
		\multirow{3}{*}{ACE-05} & Train & 17,172 & 29,006 & 4,202 & 4,664 \\
		~ & Dev & 923 & 2,451 & 450 & 560 \\
		~ & Test & 832 & 3,017 & 403 & 636 \\
		\specialrule{0em}{1pt}{1pt}
		\hline
		\specialrule{0em}{1pt}{1pt}
		\multirow{3}{*}{ERE} & Train & 14,736 & 39,501 & 6,208 & 5,054 \\
		~ & Dev & 1,209 & 3,369 & 525 & 408 \\
		~ & Test & 1,163 & 3,295 & 551 & 466 \\
		\midrule[1pt]

	\end{tabular}
	\normalsize
	\caption{Dataset statistics for supervised IE.}
	\label{tab:supervised_dataset_stats}
\end{table}
\paragraph{Baselines}
We compare the performances of fine-tuning BERT on supervised IE with the following pre-training approaches: 
1)~\textbf{\emph{BERT-Vanilla}}: we directly use the \emph{bert-base-uncased} checkpoint to fine-tune on the supervised IE tasks, which is also the same as what the baseline models do in OneIE~\cite{oneie}.
2)~\textbf{\emph{BERT-MLM}}: we initialize the BERT model with the \emph{bert-base-uncased} checkpoint and then fine-tune on the VOA corpus for 100,000 steps only using the masked language modeling loss $\mathcal{L}_{\text{MLM}}$ defined in \eqref{eqn:mlm_loss}.
3)~\textbf{\emph{BERT-SparseLT}}: our proposed approach. We pretrain the BERT model from the \emph{bert-base-uncased} checkpoint on the VOA corpus for 100,000 steps by encouraging the model to learn sparse latent types using the loss function defined in \eqref{eqn:final_loss}, with the hyper-parameters $\alpha= 0.05, \beta = 0.05, \gamma = 0.1 $. 
We did not compare our model with knowledge-enhanced pretrained language models like ERNIE~\cite{ernie} and ERICA~\cite{erica} because they use external knowledge resources aligned with the pre-training corpus, while our methods are completely self-supervised.

\paragraph{Results}
We report the F1 scores for four different IE subtasks on both ACE-2005 and ERE datasets: \emph{Entity Extraction}, \emph{Relation Extraction}, \emph{Event Detection} and \emph{Event Argument Role Labeling}, and the results are shown in Table~\ref{tab:supervised_ie_performance}.
In general, our \emph{BERT-SparseLT} model has the best performance among all model competitors, even better than the OneIE model which uses global features for fine-tuning.
In particular, our proposed method greatly improves the entity extraction performance (an absolute improvement of 7.59\% on the ERE-Entity subtask), which follows our intuition since sparse latent typing can make the model more sensitive about important entity mentions in the sentence.
We can also see that \emph{BERT-MLM} outperforms \emph{BERT-Vanilla}, which comes as no surprise since further pre-training the models on additional corpora usually leads to better performances.
The key observation that demonstrates the effectiveness of our approach is \emph{BERT-SparseLT} outperforms \emph{BERT-MLM} significantly, where our model uses exactly the same pre-training corpus without any additional information.
 %

 \begin{table*}[h]
	\centering
	\vspace{5pt}
	\small
	\begin{tabular}{c|cccc|cccc}
		\toprule[1pt] 
		Dataset & \multicolumn{4}{c|}{\textbf{\emph{ACE-2005}}} & \multicolumn{4}{c}{\textbf{\emph{ERE}}} \\
		\specialrule{0em}{1pt}{1pt}
		\cline{1-9}
		\specialrule{0em}{1pt}{1pt}
		IE subtasks  & Entity &  Trigger &  Argument & Relation
		& Entity &  Trigger &  Argument & Relation \\
		\midrule[1pt] 
		
		\emph{BERT-Vanilla} & 75.34& 66.40 &  44.90 & 41.35 & 79.39 & 55.58 & 36.76 & 31.05  \\
		\emph{OneIE-bert-base} \cite{oneie} & 77.86 & 66.91 &  48.01 & 47.88 & 79.84 & 54.33 & 37.85 & 33.39  \\
		\specialrule{0em}{1pt}{1pt}
		\hline
		\specialrule{0em}{1pt}{1pt}
		\emph{BERT-MLM} & 78.30  & 68.66 & 48.53 & 51.72 & 79.54  & 55.39& 38.35 & 33.78 \\
		\emph{BERT-SparseLT} & \textbf{81.10}  & \textbf{70.95} & \textbf{50.87} & \textbf{52.80} & \textbf{87.13} &\textbf{55.76} & \textbf{39.64} & \textbf{37.12} \\
		\midrule[1pt]
		

	\end{tabular}
		\caption{Overall test F1-scores (\%) of Supervised Joint Information Extraction.}
	\label{tab:supervised_ie_performance}
	\normalsize
	\vspace{-0.45cm}
\end{table*}

\subsection{Few-shot Named Entity Recognition}
\paragraph{Dataset} We use the most recent FewNERD~\cite{ding2021few} dataset to evaluate the performance of our proposed model on few-shot IE settings. 
The FewNERD dataset includes 8 coarse-grained and 66 fine-grained entity types, which has two experimental settings: 
1)~\emph{Inter}: The training and testing entity types are divided only based on 66 fine-grained entity types.
2)~\emph{Intra}: A more challenging setting where the training and testing entity types strictly belong to different coarse-grained entity types.
We evaluate the few-shot performance of our models in both settings and the results are shown in Table~\ref{tab:fewshot_results}.

\paragraph{Baselines and Results}
We compare our model with two competitive baselines \emph{StructShot}~\cite{structshot} and \emph{CONTaiNER-Viterbi}~\cite{das2022container}.
We replace the BERT text encoder from the state-of-the-art model \emph{CONTaiNER-Viterbi} with our text encoder pre-trained with sparse latent typing, and denote it as \emph{CONTaiNER-Viterbi + BERT-SparseLT}.
We report the F1 score of the fewshot Named Entity Recognition tasks in both \emph{intra} and \emph{inter} evaluation settings.
In general, our proposed framework greatly ourperforms previous models in both settings by 6.24\% and 3.75\% respectively while creating a new state-of-the-art of this benchmark.

 \begin{table*}[h]
	\centering
	\vspace{5pt}
	\small
	\begin{tabular}{cccccc}
		\toprule[1pt] 
		\multirow{2}{*}{\textbf{\emph{Model}} }& \multicolumn{2}{c}{\textbf{\emph{5-way}}} & \multicolumn{2}{c}{\textbf{\emph{10-way}}}&  \multirow{2}{*}{\textbf{ \emph{Average}}} \\
		\specialrule{0em}{1pt}{1pt}
		\cline{2-3}
		\cline{4-5}
		\specialrule{0em}{1pt}{1pt}
		~  & 1 $\sim$ 2 shot &  5 $\sim$ 10 shot & 
	 1 $\sim$ 2 shot &  5 $\sim$ 10 shot  \\
		\midrule[1pt] 
		\multicolumn{6}{c}{\emph{INTRA}} \\
		\specialrule{0em}{1pt}{1pt}
		\hline
		\specialrule{0em}{1pt}{1pt}
		\emph{StructShot} \cite{structshot} & 30.21 & 38.00 & 21.03 & 26.42 & 28.92\\ 
		\emph{CONTaiNER-Viterbi + BERT} \cite{das2022container} &40.40 & 53.71 &33.82&47.51 & 43.86
		\\
				\emph{CONTaiNER-Viterbi + BERT-SparseLT (Ours)}  & \textbf{47.20} & \textbf{59.67} & \textbf{40.48} & \textbf{53.04} & \textbf{50.10}
		\\
		\specialrule{0em}{1pt}{1pt}
		\hline
		\specialrule{0em}{1pt}{1pt}
		\multicolumn{6}{c}{\emph{INTER}} \\
		\specialrule{0em}{1pt}{1pt}
		\hline
		\specialrule{0em}{1pt}{1pt}
		
				\emph{StructShot} \cite{structshot} & 51.88 & 57.32 & 43.34 & 49.57 & 50.52 \\ 
		\emph{CONTaiNER-Viterbi + BERT} \cite{das2022container} &56.10 & 61.90 & 48.36 & 57.13 & 55.87
		\\
				\emph{CONTaiNER-Viterbi + BERT-SparseLT (Ours)}  & \textbf{57.14} & \textbf{66.17} & \textbf{52.75} & \textbf{62.43} & \textbf{59.62}
		\\
		
		\midrule[1pt]

	\end{tabular}
		\caption{Overall test F1-scores (\%) of Few-shot Named Entity Recognition.}
	\label{tab:fewshot_results}
	\normalsize
	\vspace{-0.15cm}

\end{table*}

\section{Analysis} \label{ANA}

In this section, we address the following research questions on Sparse Latent Typing.

\begin{figure}[htb]
  \centering
  \includegraphics[width=7cm]{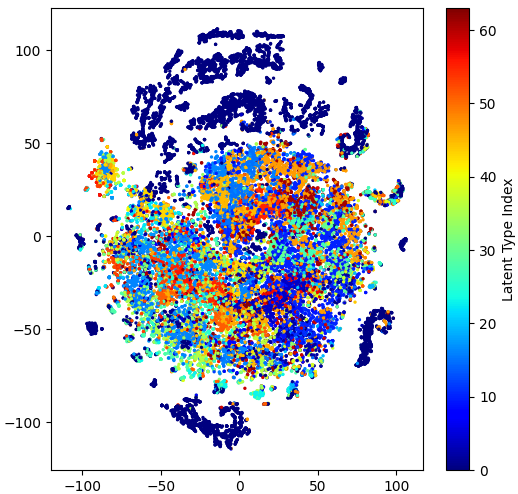}
  \caption{The t-SNE visualization of the encoded token representations  $\mathbf{x}$ with the corresponding latent type indices after 300k steps of pre-training with the sparse latent typing objective. The pre-training process is conducted on the VOA-corpus with randomly initialized parameters.}\label{TSNE}
  \vspace{-0.25cm}
\end{figure}

\paragraph{How are the latent types distributed over the encoded token representations?} We draw a t-SNE \cite{tsne} plot of the encoded token representations $\mathbf{x}$ of 1,000 sentences (30792 tokens in total) randomly sampled from the pre-training corpus in \Cref{TSNE}. The token representations are colored with their corresponding latent type indices. From the figure, we can observe that the token representations begin to be distributed as individual islands with the same colors after 300k steps of pre-training from scratch. This implies that our sparse latent typing objective can effectively encourage the clustering of the token representations in a small latent space defined by 64 randomly initialized type embeddings. We can also observe a similar trend of latent clustering for the \emph{BERT-SparseLT} model, which is illustrated in \Cref{appx:ana}, \Cref{TSNE-con}.

\paragraph{Can Typing Sparsity Loss effectively control the sparsity of token selections?} We pre-trained three \emph{BERT-SparseLT} models with different weighting factors $\beta$ of the Typing Sparsity Loss to investigate its influence on latent type selections and sentence reconstruction. (Additional samples for $\beta = 0.05$ are included in \Cref{appx:ana}, \Cref{sample}) From \Cref{sparse}, we can observe that as the $\beta$ increases the model will select fewer tokens with non-zero types. The corresponding sentence reconstruction also degenerates significantly as fewer tokens are selected by our Gumbel latent type classifier. This means that our proposed typing sparsity loss can effectively control the number of the typed tokens and thus affect the quality of reconstructed sentences. We also did the same experiments for the models pre-trained from scratch (\Cref{appx:ana}, \Cref{sparse-scr}) to illustrate that such sparse selection behavior is independent of the initialization of the model parameters.

\begin{table*}[t]
	\centering
    \small
	\begin{tabular}{ccc}
		\toprule[1pt] 
		 $\beta$ & Latent Typing of the Input Tokens & Reconstructed Sentence \\
		\midrule[1pt]
		\multirow{3}{*}{0.05} & 
		
she(58),
murdered(24),
her(1),
new(8),
york(42),

 & she was murdered in her new york office, \\
		~ & 
office(61),
just(34),
days(11),
learning(4),
waitress(62),

& just days after learning that waitress was  \\
		~ & 

accepted(60),
sundance(9),
film(63),
festival(50),

		& accepted in her the sundance film festival. \\

		\specialrule{0em}{1pt}{1pt}
		\hline
		\specialrule{0em}{1pt}{1pt}
		\multirow{3}{*}{0.06} & 		murdered(47),learning(22),

 & the government was murdered in the philippines, \\
		~ & 

waitress(42),
accepted(51),

& and a man after learning that waitress was \\
		~ & 
sundance(47),

		& accepted in the sundance festival.\\

		\specialrule{0em}{1pt}{1pt}
		\hline
		\specialrule{0em}{1pt}{1pt}		
		0.07 & 		
$\emptyset$
 & the u. s. officials have been critical of the country's.\\

\midrule[1pt]

	\end{tabular}
	\normalsize
	\caption{The latent typing and the reconstruction results of the input tokens, "She was murdered in her New York office, just days after learning that Waitress had been accepted into the Sundance Film Festival.", with different BERT-SparseLT models continually pre-trained with different values of the weighting factors $\beta$. The corresponding latent type indices for each of the tokens are noted in the parentheses. }
	\label{sparse}
\end{table*}

\begin{table*}[htbp]
	\centering
    \small
	\begin{tabular}{cc}
		\toprule[1pt] 
		 Top-10 Frequent Type Index $v$ &  Top-5 Frequent Tokens Tagged by the Type $v$ \\
		\midrule[1pt]

0 (37.86\%) & ,~(12.2\%)~ the~(12.1\%)~ .~(11.6\%)~ CLS~(8.2\%)~ SEP~(8.2\%)~\\
61 (3.82\%) & \#\#ta~(2.4\%)~ abdullah~(2.4\%)~ kabul~(2.4\%)~ afghan~(2.4\%)~ \#\#hi~(1.6\%)~\\
48 (3.29\%) & identified~(2.8\%)~ told~(2.8\%)~ ’~(1.9\%)~ \#\#j~(1.9\%)~ calls~(1.9\%)~\\
50 (2.82\%) & "~(5.5\%)~ mm~(5.5\%)~ ”~(4.4\%)~ which~(3.3\%)~ we~(2.2\%)~\\
20 (2.30\%) & but~(8.1\%)~ abortion~(8.1\%)~ "~(5.4\%)~ that~(4.1\%)~ dr~(4.1\%)~\\
53 (2.27\%) & carrier~(2.7\%)~ ka~(2.7\%)~ fraud~(2.7\%)~ claims~(2.7\%)~ harder~(2.7\%)~\\
8 (2.02\%) & also~(4.6\%)~ 2017~(3.1\%)~ among~(3.1\%)~ commission~(3.1\%)~ department~(3.1\%)~\\
38 (1.83\%) & never~(3.4\%)~ story~(3.4\%)~ afghanistan~(3.4\%)~ could~(3.4\%)~ rate~(3.4\%)~\\
62 (1.80\%) & abdullah~(5.2\%)~ indigo~(3.4\%)~ allegations~(3.4\%)~ graduating~(1.7\%)~ innovative~(1.7\%)~\\
29 (1.77\%) & low~(5.3\%)~ cost~(5.3\%)~ me~(5.3\%)~ may~(3.5\%)~ gee~(3.5\%)~\\

		\midrule[1pt]

	\end{tabular}
	\normalsize
	\caption{The frequencies of the top-10 most selected latent types $v$ and the corresponding top-5 frequent tokens tagged by the type $v$. The frequencies are computed over 100 randomly sampled sentences from the pre-training corpus and are noted in the parentheses after the tokens. The model is continually pretrained from a \emph{bert-base-uncased} checkpoint. }
	\label{freq}
    \vspace{-0.2cm}
\end{table*}

\begin{table*}[htbp]
	\centering
    \small
	\begin{tabular}{cccccccccc}
		\toprule[1pt] 
		 System &  MNLI-(m/mm) & QQP & QNLI & SST-2 & CoLA & STS-B & MRPC & RTE & \bf Average \\
		\midrule[1pt]

 $\text{BERT}_{\textit{BASE}}$-Vanilla & 84.7/84.8 & 88.5 & 91.5 & 92.4 & 58.1 & 88.7 & 90.8 & 70.4 & 83.3\\
 
 			\specialrule{0em}{1pt}{1pt}
		\hline
		\specialrule{0em}{1pt}{1pt}	
		
  $\text{BERT}_{\textit{BASE}}$-MLM-VOA  &  84.5/84.8 & 88.6 & 91.4 & 92.1 & 59.4 & 88.5 & 89.3 & 64.6 & 82.6 \\
$\text{BERT}_{\textit{BASE}}$-SparseLT-VOA  & 84.5/84.8 & 88.6 & 91.4 & 92.1 & 59.4 & 89.1 & 90.7 & 66.8  & 83.0\\
 
			\midrule[1pt]

	\end{tabular}
	\normalsize
	\vspace{-0.1cm}
	\caption{The evaluation results on the development sets of the GLUE benchmark. Following \citet{bert}, we report the F1 scores for QQP and MRPC, Spearman correlations are reported for STS-B, and
accuracy scores for the other tasks. A fixed random seed is applied for all the experiments for fair comparisons.  }
	\label{glue}
	\vspace{-0.15cm}
\end{table*}

\begin{table}[htbp]
	\centering
    \small
	\begin{tabular}{cccc}
		\toprule[1pt] 
		  $\alpha$ & $\beta$ & $\gamma$ & 5-way 1$\sim$2 shot  \\
		\midrule[1pt]
            0 & 0 & 0 & 46.10\\
            0 & 0 & 0.1 & 45.87\\
            0 & 0.05 & 0 & 45.72\\
            0.05 & 0.05 & 0 & 46.72\\
            0.05 & 0 & 0 & 46.48\\
            0.05 & 0 & 0.1 & 46.99\\
            0 & 0.05 & 0.1 & 45.94\\
            0.05 & 0.05 & 0.1 & \textbf{47.20}\\

		\midrule[1pt]

	\end{tabular}
	\normalsize
	\vspace{-0.1cm}
	\caption{The test F1 scores on the \emph{INTRA} 5-way 1$\sim$2 shot setting of the FewNERD dataset for different  \emph{BERT-SparseLT} models continually pre-trained with different loss weighting factors, $\alpha$, $\beta$, $\gamma$. }
	\vspace{-0.25cm}
	\label{abl}
\end{table}

\paragraph{To what extent are the learned latent types interpretable?} In \Cref{freq}, we calculate the frequencies for the top-10 most selected latent types $v$ and the probability $P(x |z = v)$ of the top-5 frequent tokens tagged by the type $v$. The statistics are computed over 100 randomly sampled sentences from the pre-training VOA corpus with the \emph{BERT-SparseLT} model. We can observe that the zero type (index 0) is mostly associated with less meaningful tokens such as ",", "CLS","SEP", which are used for delimiting the semantics. This means that the model can effectively learn to select more meaningful tokens through our sparse latent typing objective. We can also observe that the type 61 seems more correlated with the physical locations and the type 50 is mostly related to functional words. We also do the same analysis for the model pre-trained from scratch in  \Cref{appx:ana}, \Cref{freq-scr}. The results appear to be more interpretable due to more consistent training dynamics than continual pre-training from a checkpoint.

\paragraph{How could different loss combinations affect the model performance?} We conduct an ablation study of the loss weighting factors of the \emph{BERT-SparseLT} model to illustrate the influence of different loss combinations on the few-shot NER performance in \Cref{abl}. Including all the four losses produces the best test performance on the 5-way 1$\sim$2 shot evaluation of the \emph{Intra} setting of the FewNERD benchmark. This proves that all the training objectives are necessary for improving the generalizability of the token representations learned by the encoder. We also include the reconstruction and the latent typing results of the models trained with $ \alpha = 0.05 $ in \Cref{appx:ana}, \Cref{abl_tab} for further qualitative analyses.

\paragraph{Can sparse latent typing improve the sentence-level Natural Language Understanding (NLU) ability?} We evaluate \emph{BERT-Vanilla}, \emph{BERT-MLM} and the \emph{BERT-SparseLT} models on the General Language
Understanding Evaluation (GLUE) benchmark to demonstrate the influence of sparse latent typing on NLU, and the results are shown in \Cref{glue}. The finetuning hyperparameters are shown in \Cref{appx:ana}, \Cref{hyp_glue}. Our \emph{BERT-SparseLT} model obtains slightly worse results than the vanilla BERT, but still has marginal improvement over the MLM baseline that excludes the sparse latent typing objectives. The inferior results are as expected for two reasons: 1) The evaluation of the GLUE benchmark heavily relies on the [CLS] token which is always latent typed as a zero-type by the \emph{BERT-SparseLT} model  and thus lacks the enough training signals for fine-grained clustering in the latent space. 2) The VOA corpus for continual pretraining is in the specific news domain, which may not be beneficial for general NLU. We hypothesize that large-scale pretraining from scratch of an encoder-decoder model should overcome these limitations, and we leave it as the future work due to the limitation of the computation resources.

\section{Conclusion}
In this paper, we propose a novel language model pre-training framework that encourages the model to sparsely extract sentence-level keywords with meaningful latent types in a completely self-supervised manner. 
Experimental results and analysis demonstrate that incorporating sparse latent type learning early in the pre-training stage will not only facilitate the model to learn sentence-level keyword selections with interpretable latent types, but also improves downstream Information Extraction tasks in both supervised and few-shot settings.
\section{Limitations}
One primary limitation of our framework is that the language model pretrained with sparse latent typing might only improve performance on Information Extraction tasks.
Although this is intuitive since IE shares essential similarity with latent typing, it is exciting to see whether our model can improve other downstream tasks such as natural language generation and abstractive summarization.

Another limitation of our work is that, due to the lack of the computation resources, we did not conduct experiments of large-scale pretraining from scratch for a comprehensive examination of our framework's ability on improving the general NLU performance. For future works, it is also worth exploring on whether the sparse latent typing objective can improve the machine translation performance by regularizing a sparse and unified latent space for cross-lingual meaning representations.

Finally, our model is only capable of extracting sentence-level keywords with latent types, but is not designed to learn a comprehensive graph structure for each input sentence.
Although it is debatable whether a more complex latent graph representation is better than concise latent types, it is still worth adding this into future work plans.

\section*{Acknowledgement}
We thank the anonymous reviewers helpful suggestions. This material is based upon work supported in part by the IBM-Illinois Discovery Accelerator Institute and by the National Science Foundation
under Grant No. 1801652. 
This research is also based upon work supported by U.S. DARPA KAIROS Program No. FA8750-19-2-1004 and U.S. DARPA AIDA Program No. FA8750-18-2-0014. The views and conclusions contained herein are those of the authors and should not be interpreted as necessarily representing the official policies, either expressed or implied, of DARPA, or the U.S. Government. The U.S. Government is authorized to reproduce and distribute reprints for governmental purposes notwithstanding any copyright annotation therein.
We also thank Suyu Ge for early discussions on this project.

\bibliographystyle{acl_natbib}
\bibliography{anthology,custom}

\appendix
\section{Implementation Details}\label{appx:implementation}
Our transformer decoder layer follows the same architecture as the BART model \cite{bart}. The word embeddings for both the text encoder and decoder are tied together. Our implementation is based on the Transformers codebase \footnote{\url{https://github.com/huggingface/transformers}}. \cite{wolf-etal-2020-transformers}.

We train our model on 4 NVIDIA Tesla V100 GPUs with 16GB memory. The pre-training time for \emph{bert-base-uncased} model on the VOA-corpus is about 12 hours. A linear learning rate scheduling with warm-up is adopted. For Gumbel latent typing, we adopt the following  temperature annealing schedule,
$$
\tau = \max(5\times 0.99997^T, 0.5)
$$
where T is the number of training steps.  We continually pretrain a BERT-base model from the \emph{bert-base-uncased} checkpoint for 100k steps. For the experiment of pre-training from scratch, we only train a RoBERTa-large \cite{roberta} model on the VOA-corpus for 300k steps, given the limitation of our computation resources. The detailed hyper-parameters are summarized in \Cref{hyp2} and \Cref{hyp1}.

For fine-tuning on downstream tasks, we replace the BERT model used in the state-of-the-arts with our pre-trained BERT-SparseLT model and follow the same hyper-parameter settings. Specifically, for supervised IE, we use the codebase from OneIE \cite{oneie} \footnote{\url{http://blender.cs.illinois.edu/software/oneie/}}, and for few-shot IE, the codebase from CONTaiNER \cite{das2022container} is adopted.\footnote{\url{https://github.com/psunlpgroup/CONTaiNER}}.

\begin{table}[htbp]
	\centering
	\small
	\begin{tabular}{cc}
		\toprule[1pt] 
		Hyper-parameters & Values\\
		\midrule[1pt]
		 Weighting factor for Reconstruction loss, $\alpha$ & 0.05 \\
		Weighting factor for Typing Sparsity loss, $\beta$ & 0.05 \\
		 Weighting factor for KL-divergence, $\gamma$ & 0.1 \\

		\specialrule{0em}{1pt}{1pt}
		\hline
		\specialrule{0em}{1pt}{1pt}
		Model dimension, $d_m$ & 768 \\
		Number of latent types, $V_c$ & 64 \\
		\specialrule{0em}{1pt}{1pt}
		\hline
		\specialrule{0em}{1pt}{1pt}
		 Batch size  & 32 \\
		 Learning rate warm-up steps & 300 \\
		 Max trainig steps & 100,000 \\
		 Adam $\beta_1$ & 0.9 \\
		 Adam $\beta_2$ & 0.999  \\
		 Adam $\epsilon$ & 1e-8 \\
		 Weight decay  & 0.01 \\
		 Learning rate & 1e-5 \\
		 Gradient clipping norm & 1.0 \\
         MLM masking probability & 0.15 \\
        \specialrule{0em}{1pt}{1pt}
		\hline
        \specialrule{0em}{1pt}{1pt}
        Number of decoder layers & 1 \\
        Dropout rate & 0.1 \\
        Activation function & \emph{GELU} \\

		\midrule[1pt]
	\end{tabular}
	\normalsize
	\vspace{-0.2cm}
	\caption{Detailed settings for model hyper-parameters when pretraining from a \emph{bert-base-uncased} checkpoint. }
	\vspace{-0.2cm}
	\label{hyp2}
\end{table}

\begin{table}[htbp]
	\centering
	\small
	\begin{tabular}{cc}
		\toprule[1pt] 
		Hyper-parameters & Values\\
		\midrule[1pt]
		 Learning Rate & \{1e-5, 2e-5, 5e-5\}\\
		 Batch size  & 32 \\
		 Learning rate warm-up ratio & 0.06 \\
		 Max trainig epochs & 10 \\
	
		 Weight decay  & 0.1 \\
		 Learning rate decay & Linear \\

		\midrule[1pt]
	\end{tabular}
	\normalsize
	\vspace{-0.2cm}
	\caption{Hyperparameters for finetuning the \emph{BERT-SparseLT} model on the GLUE benchmark. We follow the settings in \citet{roberta}. }
	\vspace{-0.2cm}
	\label{hyp_glue}
\end{table}

\begin{table}[htbp]
	\centering
	\small
	\begin{tabular}{cc}
		\toprule[1pt] 
		Hyper-parameters & Values\\
		\midrule[1pt]
		 Weighting factor for Reconstruction loss, $\alpha$ & 1.3 \\
		Weighting factor for Typing Sparsity loss, $\beta$ & 0.2 \\
		 Weighting factor for KL-divergence, $\gamma$ & 0.1 \\

		\specialrule{0em}{1pt}{1pt}
		\hline
		\specialrule{0em}{1pt}{1pt}
		Model dimension, $d_m$ & 1024 \\
		Number of latent types, $V_c$ & 64 \\
		\specialrule{0em}{1pt}{1pt}
		\hline
		\specialrule{0em}{1pt}{1pt}
		 Batch size  & 32 \\
		 Learning rate warm-up steps & 300 \\
		 Max trainig steps & 300,000 \\
		 Adam $\beta_1$ & 0.9 \\
		 Adam $\beta_2$ & 0.999  \\
		 Adam $\epsilon$ & 1e-8 \\
		 Weight decay  & 0.01 \\
		 Learning rate & 1e-5 \\
		 Gradient clipping norm & 1.0 \\
         MLM masking probability & 0.15 \\
        \specialrule{0em}{1pt}{1pt}
		\hline
        \specialrule{0em}{1pt}{1pt}
        Number of decoder layers & 1 \\
        Dropout rate & 0.1 \\
        Activation function & \emph{GELU} \\

		\midrule[1pt]
	\end{tabular}
	\normalsize
	\vspace{-0.2cm}
	\caption{Detailed settings for model hyper-parameters when pretraining from scratch.}
	\vspace{-1.0cm}
	\label{hyp1}
\end{table}

\begin{figure}[h]
  \centering
  \includegraphics[width=7cm]{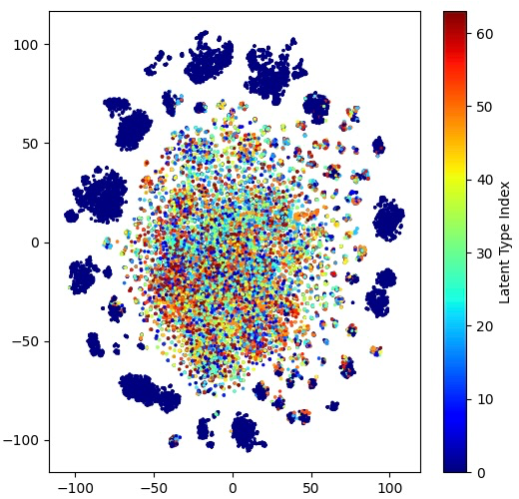}
  \caption{The t-SNE visualization of the encoded token representations  $\mathbf{x}$ with the corresponding latent type indices for the \emph{BERT-SparseLT} model. The model is continually pretrained from a \emph{bert-base-uncased} checkpoint.}\label{TSNE-con}
\end{figure}

\section{Additional Analysis}\label{appx:ana}
This section includes all the additional figures and the tables mentioned in \cref{ANA}.

\begin{table*}[t]
	\centering
    \small
	\begin{tabular}{ccc}
		\toprule[1pt] 
		 $\beta$ & Latent Typing of the Input Tokens & Reconstructed Sentence \\
		\midrule[1pt]
		\multirow{3}{*}{0.2} & 
She(36),
murdered(20),
her(57),
New(17),
office(1),
just(51),
 & She said murdered of her New York office,\\
		~ & 
days(33),
after(23),
learning(62),
that(39),
Wait(10),
ress(10),
had(59),
&just days after learning that wanderingress had \\
		~ & 

accepted(11),
into(59),
Sund(10),
ance(10),
Film(10),
Festival(10)
		&been accepted into the Sundance Film Festival. \\

		\specialrule{0em}{1pt}{1pt}
		\hline
		\specialrule{0em}{1pt}{1pt}
		\multirow{3}{*}{0.3} & 		murdered(59),
her(63),
office(7),
just(63),
 & The U.S. officials have just just days of learning\\
		~ & 
days(49),
learning(49),
Wait(7),
ress(43),
had(24),
& the countryress had been accepted had been  \\
		~ & 
accepted(24),
Sund(43),
ance(43),
Film(43),
Festival(43)
		& a Sundance Film Festival, the\\

		\specialrule{0em}{1pt}{1pt}
		\hline
		\specialrule{0em}{1pt}{1pt}		
		\multirow{3}{*}{0.4} & 		murdered(9),

 & The U.S. President Barack Obama, the U.S. \\
		~ & 
learning(63),
Wait(9),
ress(50),

&  akaress, and the accepted the Sundance \\
		~ & 
accepted(63),
Sund(63),
ance(63),
Film(9),
Festival(9)
		&  Film Festival.\\

\midrule[1pt]

	\end{tabular}
	\normalsize
	\caption{The latent typing and the reconstruction results of the input tokens, "She was murdered in her New York office, just days after learning that Waitress had been accepted into the Sundance Film Festival.", with different models pre-trained from scratch with different values of the weighting factors $\beta$. The corresponding latent type indices for each of the tokens are noted in the parentheses. }
	\label{sparse-scr}
\end{table*}

\begin{table*}[htbp]
	\centering
    \small
	\begin{tabular}{cc}
		\toprule[1pt] 
		 Top-10 Frequent Type Index $v$ &  Top-5 Frequent Tokens Tagged by the Type $v$ \\
		\midrule[1pt]

0 (30.55\%) & ,~(14.1\%)~ the~(13.8\%)~ .~(13.1\%)~ CLS~(10.3\%)~ SEP~(10.3\%)~\\
47 (5.86\%) & in~(3.2\%)~ and~(3.2\%)~ said~(1.6\%)~ as~(1.1\%)~ out~(1.1\%)~\\
10 (5.73\%)& abortion~(2.7\%)~ cost~(2.2\%)~ Indigo~(1.6\%)~ eta~(1.6\%)~ innovative~(1.1\%)~\\
16 (5.14\%) & eta~(1.8\%)~ parallel~(1.8\%)~ autism~(1.8\%)~ MMR~(1.8\%)~ Smart~(1.2\%)~\\
61 (3.2\%) & that~(7.8\%)~ was~(5.9\%)~ a~(3.9\%)~ has~(2.9\%)~ would~(2.9\%)~\\
46 (2.76\%)& Abdullah~(5.7\%)~ Ge~(2.3\%)~ Jones~(2.3\%)~ measles~(2.3\%)~ Minnesota~(2.3\%)~\\
36 (2.73\%)& was~(3.4\%)~ that~(3.4\%)~ on~(3.4\%)~ )~(2.3\%)~ she~(2.3\%)~\\
15 (2.70\%)& ,"~(3.5\%)~ told~(2.3\%)~ could~(2.3\%)~ statistics~(2.3\%)~ (~(2.3\%)~\\
45 (2.54\%)& and~(3.7\%)~ under~(2.5\%)~ Afghan~(2.5\%)~ new~(2.5\%)~ vaccine~(2.5\%)~\\
2 (2.23\%)& cost~(2.8\%)~ .,~(1.4\%)~ hopes~(1.4\%)~ challenges~(1.4\%)~ os~(1.4\%)~\\

\midrule[1pt]

	\end{tabular}
	\normalsize
	\caption{The frequencies of the top-10 most selected latent types $v$ and the corresponding top-5 frequent tokens tagged by the type $v$. The frequencies are computed over 100 randomly sampled sentences from the pre-training corpus and are noted in the parentheses after the tokens. The model is pretrained from scratch on the VOA corpus. We can observe that the zero type (index 0) is mostly associated with less meaningful tokens. We can also observe that the type 46 seems more correlated with the physical locations and the types 47 and 36 are mostly related to functional words. Type 16 is also meaningful as it appears to be related to the discussion of the potential linking of MMR vaccines to autism in children.}
	\label{freq-scr}
\end{table*}

\begin{table*}[htbp]
	\centering
    \small
	\begin{tabular}{p{0.03\linewidth} p{0.03\linewidth} p{0.03\linewidth} p{0.35\linewidth} p{0.35\linewidth}}
		\toprule[1pt] 
		  $\alpha$ & $\beta$ & $\gamma$ & Latent Typing & Reconstructed Sentence  \\
		\midrule[1pt]

0.05 & 0.05 & 0.1 &

she(58), murdered(24), her(1), new(8), york(42), office(61), just(34), days(11), learning(4), waitress(62), accepted(60), sundance(9), film(63), festival(50) &

she was murdered in her new york office, just days after learning that waitress was accepted in her the sundance film festival. \\                                    

		\specialrule{0em}{1pt}{1pt}
		\hline
		\specialrule{0em}{1pt}{1pt}
0.05 & 0 & 0.1 &

CLS(46), she(54), was(13), murdered(35), in(47), her(28), new(28), york(32), office(25), ,(26), just(46), days(14), after(31), learning(44), that(4), waitress(44), had(41), been(27), accepted(58), into(41), the(30), sundance(61), film(5), festival(36), .(1) &

she was murdered in her new york office, just days after learning that waitress had been accepted into the sundance film festival. \\

		\specialrule{0em}{1pt}{1pt}
		\hline
		\specialrule{0em}{1pt}{1pt}

0.05 & 0.05 & 0 & $\emptyset$ &

the u. s. military chief said. \\

		\specialrule{0em}{1pt}{1pt}
		\hline
		\specialrule{0em}{1pt}{1pt}
		
0.05 & 0.0 & 0 & 

CLS(38), she(4), was(4), murdered(4), in(4), her(4), new(4), york(4), office(4), ,(4), just(4), days(4), after(4), learning(4), that(4), waitress(4), had(4), been(4), accepted(4), into(4), the(4), sundance(4), film(4), festival(4), .(4), SEP(48) &

she was murdered in her new york office, just days after learning that waitress had been accepted into the sundance film festival. \\

\midrule[1pt]

	\end{tabular}
	\normalsize
	\caption{The latent typing and the reconstruction results of various \emph{BERT-SparseLT} models continually pretrained with different $\alpha$, $\beta$ and $\gamma$ values. The input sentence is "She was murdered in her New York office, just days after learning that Waitress had been accepted into the Sundance Film Festival.". From the table, we can see that removing the typing sparsity loss ($\beta =0$) will result the model to select all the input tokens, and removing the KL-divergence term ($\gamma =0$) will cause the model to either not select any tokens (assigning all the tokens as the zero type) or have almost the same latent type (e.g. type 4) for all the input tokens. The corresponding latent type indices for each of the tokens are noted in the parentheses.} 
	\label{abl_tab}
\end{table*}

\begin{table*}[htbp]
	\centering
    \small
	\begin{tabular}{p{0.3\linewidth} p{0.3\linewidth} p{0.3\linewidth}}
		\toprule[1pt] 
		  Input Tokens & Latent Typing & Reconstructed Sentence  \\
		\midrule[1pt]
		\multicolumn{3}{c}{\emph{In-domain Sentences}} \\
		\specialrule{0em}{1pt}{1pt}
		\hline
		\specialrule{0em}{1pt}{1pt}
A 45-year-old man who was tackled down and arrested by police after allegedly attacking officers with a knife in South Roxana, Illinois, has been charged by the local prosecutors. &
45(36), year(48), old(16), man(35), who(8), was(52), tackled(37), down(59), arrested(19), police(36), after(28), allegedly(14), attacking(48), officers(60), with(5), knife(7), sou
th(4), ro(62), \#\#xa(48), \#\#na(48), illinois(20), has(50), charged(49), local(39), prosecutors(38), &                                                
a 45 - year - old man who was tackled down and arrested by police after allegedly attacking officers with a knife in south roxana, illinois, has been charged at the local prosecutors. \\

			\specialrule{0em}{1pt}{1pt}
		\hline
		\specialrule{0em}{1pt}{1pt}

Natural language processing (NLP) is a subfield of linguistics, computer science, and artificial intelligence concerned with the interactions between computers and human language,
 in particular how to program computers to process and analyze large amounts of natural language data. &      
natural(20), language(24), processing(54), nl(29), \#\#p(42), )(43), sub(34), \#\#field(61), linguistics(29), computer(29), science(15), artificial(59), intelligence(59), concerned(12), interactions(19), between(34), computers(20), human(28), language(21), particular(15), how(38), program(6), computers(61), process(32), analyze(13), large(53), amounts(48), natural(25), language(10), data(21), &
natural language processing ( nlp ) is a subfield of linguistics, computer science and artificial intelligence concerned about the interactions between computers and human language, in particular how to program computers to process to process and analyze large amounts of natural language data. \\

			\specialrule{0em}{1pt}{1pt}
		\hline
		\specialrule{0em}{1pt}{1pt}	

		\multicolumn{3}{c}{\emph{Out-of-domain Sentences}} \\
		\specialrule{0em}{1pt}{1pt}
		\hline
		\specialrule{0em}{1pt}{1pt}
		
The objective of sparse latent typing is to find pairs of latent types and token representations that are as compact as possible but still contain the necessary information for reconstructing the original input sentences.
& objective(30), sparse(27), late(48), 
\#\#nt(42), typing(23), find(60), pairs(63), late(29), \#\#nt(4), types(53), token(53), representations(6), as(49), compact(59), as(58), possible(20), but(58), still(20), contain(48), necessary(20), information(35), rec(36), \#\#ons(42), \#\#tructing(44), original(6), input(27), sentences(44) & the objective of sparse latent typing is to find pairs of latent types and token representations, as compact as possible but still contain the necessary information for reconstructing the original input sentences.\\

			\specialrule{0em}{1pt}{1pt}
		\hline
		\specialrule{0em}{1pt}{1pt}	
		
Our approach provides the decoder model with a shortcut to directly access the encoded token representations, so that the latent representation for each of the input tokens can be learned as an auxiliary type representation. & our(20), approach(20), provides(48), deco(19), \#\#der(13), model(27), with(16), short(49), \#\#cut(61), directly(18), access(48), encoded(25), token(53), representations(6), so(2), that(59), late(49), \#\#nt(4), representation(22), each(26), input(25), token(53), can(41), learned(38), as(58), auxiliary(32), type(30), representation(53)  & our approach provides the decoder model with a shortcut to directly access the encoded token representations, so that the latent representation of each of the input tokens can be learned as an auxiliary type representation representation. \\

\midrule[1pt]

	\end{tabular}
	\normalsize
	\caption{Sample latent typing and sentence reconstruction results of the continually pretrained \emph{BERT-SparseLT} model for both the in-domain and the out-of-domain sentences. The in-domain sentences are sampled from the VOA corpus and the Wikipedia, while the out-of-domain sentences are from the main content of this paper. }
	\label{sample}
\end{table*}

\end{document}